# Real-time Tracking Based on Neuromrophic Vision

Hongmin Li

Center for Brain-Inspired Computing Research (CBICR), Optical Memory National Engineering Research Center, Department of Precision Instrument, Tsinghua University, Beijing 100084, *China*

Jing Pei, and Guoqi Li

Center for Brain-Inspired Computing Research (CBICR), Optical Memory National Engineering Research Center, Department of Precision Instrument, Tsinghua University, Beijing 100084, *China*

*Abstract*—Real-time tracking is an important problem in computer vision in which most methods are based on the conventional cameras. Neuromorphic vision is a concept defined by incorporating neuromorphic vision sensors such as silicon retinas in vision processing system. With the development of the silicon technology, asynchronous event-based silicon retinas that mimic neuro-biological architectures has been developed in recent years. In this work, we combine the vision tracking algorithm of computer vision with the information encoding mechanism of event-based sensors which is inspired from the neural rate coding mechanism. The real-time tracking of single object with the advantage of high speed of 100 time bins per second is successfully realized. Our method demonstrates that the computer vision methods could be used for the neuromorphic vision processing and we can realize fast real-time tracking using neuromorphic vision sensors compare to the conventional camera.

## I.  INTRODUCTION

Real-time tracking of the motion of objects remains the heart of many application problems in computer vision[1-6], especially in motor control of robots[1, 4]. Despite that a number of tracking algorithms have been developed and got successful applications in many aspects, they are all based on conventional frame-based cameras which suffer from huge bandwidth requirements, more data storage and low time resolution with low frame rates. Neuromorphic vision[7, 8] is inspired from the biological retina and aims to develop a vision processing mechanism by incorporating neuromorphic vision sensors such as silicon retinas in vision tasks. Artificial silicon retina is a device that encodes the natural images or videos to neural spike patterns. The modern advanced silicon technology has resulted in the appearance of many kinds of silicon retinas. Neuromorphic vision based on silicon retinas has the advantages of less power, less data storage and less computational requirements than the conventional frame-based vision processing system[9].

To address the real-time tracking problem, many works based on asynchronous event-based sensors[10] have been proposed. In contrast to the traditional frame-based cameras its output is a continuous stream of spike-like asynchronous temporal events similar to the biological retinas. The spikes of the artificial silicon retina is encoded in Address Event Representation (AER)[7, 11] that integrates the event's location, time, and polarity. The AER is an event-driven communication technology widely used in neuromorphic systems. Every pixel of the sensors responses independently to the light intensity changes by generating AER's spikes every one of which is with microsecond-precise timing. When a silicon retina moves, the pixels at the intensity edges will trigger spikes. It's not the image gradients but only the locations, times and polarities of brightness changes are measured. With high time resolution, the silicon retinas will not suffer from motion blur, which is a severe problem for the traditional frame-based cameras in traditional feature tracking applications.

There have been many applications of object tracking using synchronous event-based sensors[12-18]. Existing methods are mainly based on clustering the spikes to find the moving features that are then tracked quickly. J. Conradt et al.[13] developed a neuromorphic vision system to balance a normal pencil on its tip by fast detecting the position changes of the pencil using two silicon retinas. M. Litzenberger et al. [17] developed an embedded vision system which is used in vehicle tracking for a traffic-monitoring in real time. T. Delbruck et al.[14] developed a hybrid neuromorphic vision system which consisted of a silicon retina, a computer, and a servo motor to achieve a fast sensory-motor controller based on cluster tracking algorithm. Z. Ni et al.[18] presented a high speed vision system using an asynchronous address-event sensor to detect the Brownian motion and adapted the method to the microrobotic systems fulfilling an extremely fast vision feedback. In fact, the cluster tracking algorithm cannot implement the true object tracking as the computer vision has always been doing. It just regards the mass of spikes as an object, implements the position detection, and doesn't consider the appearance of the objects. In other words, the cluster tracking algorithm cannot recognize the pattern of objects it is tracking. Similarly, Z. Ni et al.[19] proposed a method based on a time-coded, frame-free visual data to implement a continuous and iterative estimation of the geometric transformation between the preconstructed model and the events representing the tracked object. It updates the model at the arriving of every event, which is easily influenced by the noises.

Different from the above, in this work we combined the AER spike patterns with the neural coding mechanisms, and found that the spike count coding [20] in a time bin can be used to distinguish the patterns of the moving scene. Then we divide all the moving process into many time bins. In every time bin, we decode the spike trains of the pixel population using spike count coding mechanism. The number of the spikes of every pixel represents the information of the scene in this time window. We put the series of the time bins into a tracking algorithm based on compressive sensing[5] and realized real-time compressive tracking of an object with a high time resolution, high accuracy and robustness. We connected a silicon retina to a PC with Windows 7 system and recorded the spike streams of the moving objects with the open source software named jAER[21] developed in Java. We ran the algorithm routine of the real-time compressive tracking on a PC with Windows 7 system. The silicon retina, PC, jAER and algorithm routine constitute a neuromorphic vision real-time tracking system. In the following sections, we will analyze the system units one by one.

## II. EXPERIMENT AND ALGORITHM

The neuromorphic vision system contains hardware part and software part as shown in Figure 1. The key component of the hardware part is the Dynamic Vision Sensor (DVS)[10] which is inspired by the transient vision pathway of the biological retina. The DVS responses to the moving scenes and generates spike trains in every pixel element similar to the output cells of the biological vision system. We researched the neural coding mechanism in computational neuroscience and proposed a spike count decoder for the DVS output spike trains.

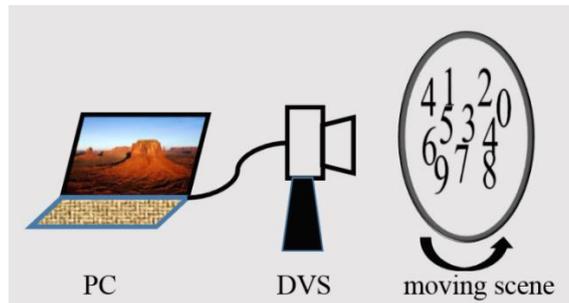

Figure.1. Main components of our neuromorphic vision real-time tracking system.

### A. Silicon Retina Sensor

In this section, the properties of the silicon retina sensor are summarized. Traditional cameras see the world as a series of frames. Vast quantities of redundant information exists in the successive frames. In addition, low frame rate limits the speed and accuracy of real-time object tracking. Finally, the frame-based cameras have high power requirements. DVS is the first commercial artificial silicon retina belonging to a neuromorphic vision sensor class with a resolution of $128 \times 128$ pixels. Unlike the traditional CCD or CMOS cameras, DVS contains an array of autonomous, self-signaling pixels that independently respond to the temporal changes in light intensity and place their address on an asynchronous arbitrated bus. Each pixel contains an active continuous-time front-end logarithmic photoreceptor and a self-time switched-capacitor differencing circuit. The sensor has an array mismatch of 2.1% in relative intensity spike threshold. The dynamic range of the sensor is >120 dB, while the chip power consumption is 23 mW. The minimum latency is about 15 us, which represents an effective single pixel bandwidth of 66 kHz. The event threshold of about 10% contrast can be set, which allows the sensor to sense real-world contrast signals. Because of the asynchronous response property, the spikes have a high timing precision of the pixel response with an "effective frame rate" of typically several kHz.

The DVS is event-driven instead of clock-driven, which the static scenes will not trigger spikes. When the scene changes, it will produces output. The change includes a modulated light source, for example, a blinking LED, or moving scenes relative to DVS. As a result, the DVS performs on-chip data compression which avoids a great deal of redundant information. Physically, the photons hit the photodiode and produce a photocurrent, which is then converted into a voltage. When the difference between the current voltage and the last sampled voltage described by TCON is larger than a certain threshold described by $\theta$, a spike is sent off chip. The temporal contrast denoted as TCON which the pixel is sensitive to is defined as

$$\text{TCON} = \frac{1}{I(t)} \frac{dI(t)}{dt} = \frac{d(\ln(I(t)))}{dt} \tag{1}$$

where $I(t)$ is the photocurrent which is a function of time.

The spikes of the sensor are outputted in the form of AER. We represent the each spike by a tuple ((x, y), t, p). The coordinates (x, y) identify the location of the pixel that triggered the spike. The scalar *t* is the timestamp of the spike, which has a 10 us resolution. The value *p* is the polarity of the spike. The positive change in light intensity generates a +1 polarity, while the negative change a -1 polarity.

Figure.2 illustrates how the +1 and -1 spikes of single pixel are internally generated and output in response to an input signal. In Figure.2 (a), the moving scene with a linear change of light intensity in a time window is detected by a pixel which outputs four spikes at four different time points. In Figure.2 (b), the change of light intensity is a little more complicated with a nonlinear form. The pixel that detects the brightness change outputs six spikes at six different time points. The rate of generated spikes can be approximated with

$$f(t) = \text{spike rate}(t) \approx \frac{TCON(t)}{\theta} \quad (2)$$

where $\theta$ is the relative change threshold.

Finally, the tracking system we built has a USB2.0 interface that transmits the spikes to a host PC. Then we can develop algorithms very conveniently for using the silicon retina output and process many kinds of real-world vision tasks based on these effective neuromorphic vision systems.

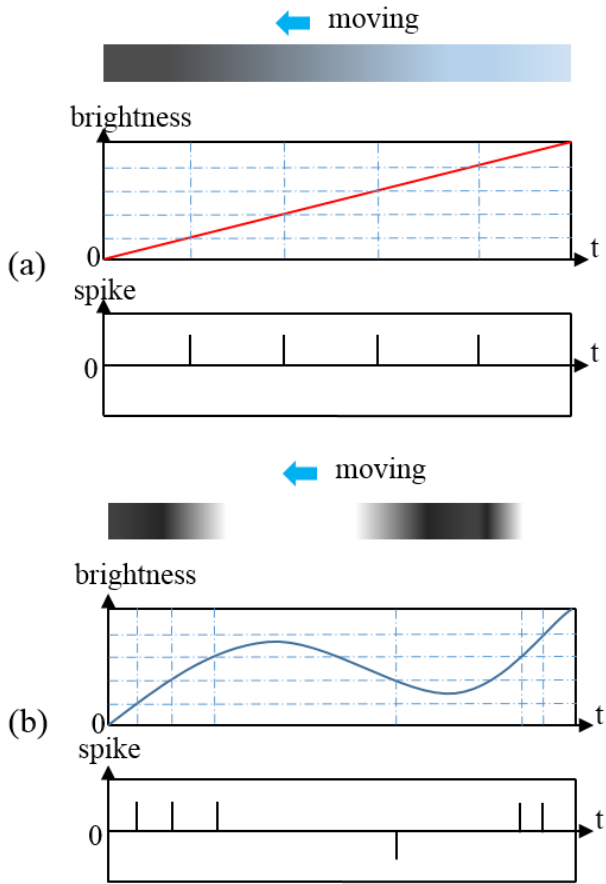

Figure.2. Principle of how the +1 and -1 spikes are internally represented and output in response to the changes of light intensity. (a) The moving scene projects a linear brightness change to a pixel of DVS, and triggers four spikes at four different time points. b) Spike train of the pixel. (b)The moving scene projects a nonlinear brightness change to a pixel of DVS, and triggers six spikes at six different time points.

*B. Spke count coding*

As the DVS output a series of spike trains similar to the output cells of biological retinas, we combine the neural coding mechanisms with the silicon retina spikes. The neural code is a language applied to the nervous system. To understand how spikes represent the symbols remains the fundamental problem of neural coding. Not only in biological nerve system, but also in the neuromorphic engineering, the spikes are stereotyped events precisely defined in time, and the amplitude and pulse width of spikes contain no information. There are three main classes of neural coding mechanisms containing rate coding, temporal coding and population coding. Spike count coding belongs to rate coding, which is obtained by counting the amount of the spikes in a time bin. In very time bin, the number of spikes is counted as the information representation of the stimulus in neural processing system. The assembly of all the neurons make up a population coding. In Figure.3, the spike count coding of three stimulus is shown. Five neurons is used to response to the three signals. The stimulus1 is encoded in the time window with the length of T and represented as a vector of <5, 1, 3, 2, 3>. Then the stimulus2 and stimulus3 are <4, 2, 2, 3, 1> and <1, 2, 2, 2, 1> in two time windows with the same length, respectively. Although a lot of detail such as the changing process of the brightness encoded in the timing of the spikes is lost, different stimuli can be effectively distinguished with a population of neurons. In the real-time tracking, change of brightness on single pixel is meaningless. The mass of the spikes triggered by plenty of pixels form the shape of the moving scene which is also known as the population coding. As a conclusion, we can say that spike count coding and population coding can be combined to describe the appearance of the objects. Based on the appearance, many real-time vision tracking algorithms can be developed to realize fast, efficient and robust object tracking tasks.

In fact, the spike count coding is extremely fit for the output of the silicon retinas. Each pixel of the sensor is regarded as an output neuron of the biological retinas. The pixels response to the visual stimulus and encode the change of light intensity to spikes. For a dynamic signal input, each pixel detects the brightness contrast in real time and output a continuous event stream. As a result, the number of the spikes in a time bin represents the total change of the light intensity. The silicon retinas only output spikes in an asynchronous and continuous way, which achieves a high time resolution and eliminates redundancy. For a moving scene, the spike count coding of a pixel encodes the texture in a small distance that travels in a small time bin. The pixel array of the sensor encodes the total texture information of the moving scene as a population coding similar to the neuronal population. We determine the length of every time bin reasonably to ensure that the total spikes in the time bin pile up in the tracking objects in the two-dimensional plane. Figure.3 shows the spike counting coding process of the moving scenes. We print the digit '3' and digit '5' in Times New Roman font on two A4 papers and move the papers in front of the DVS up, down, left and right as is shown in the first row. The DVS detects the moving edges of the digits and output a stream of asynchronous and temporal events as is shown in the second row. We draw all the event points on a two-dimensional plane, which is shown in the third row. Finally, we count the number of the spikes of every pixels in the time bin, and represent the numbers using different colors as is shown in the fourth row. In out experiment, the length of the time bin is chosen as 10 ms, which means the speed of the tracking is 100 step per second.

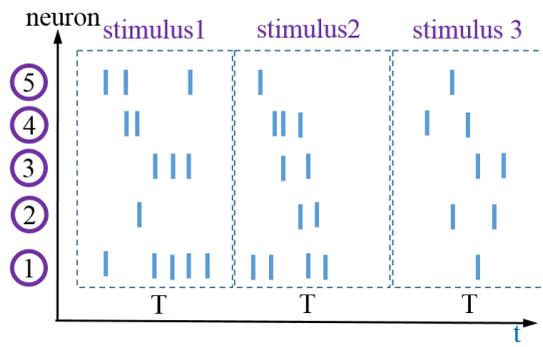

Figure.3. Spike count coding.

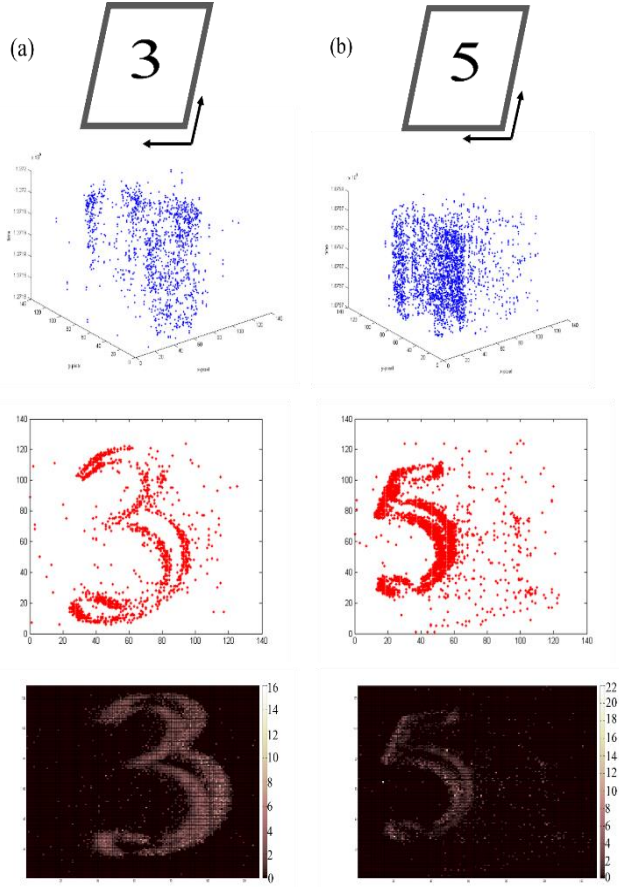

Figure.4. DVS spikes and spike count coding of two moving pictures of (a): digit '3' and (b): digit '5'. The first row is the scenes of while papers on which we draw the digit '3' and digit '5', respectively. The second row is output events drawn in three-dimensional space of *(x, y, t)*, *(x, y)* is the pixel coordinates, *t* is the time axis. The third row is the output events drawn in two-dimensional space of *(x, y)*. The fourth row is the spike count coding map of $128 \times 128$ pixels.

*C. Tracking Algorithm*

In this section, we summarize the compressive tracking algorithm we used in the neuromorphic vision tracking problem based on the appearance of objects in spike count coding as is shown in Figure.5. Different from other methods, compressive tracking algorithm contains an appearance model based on features extracted from the multi-scale image feature space and a very sparse measurement matrix to compress samples of foreground targets and the background. It runs in real-time way and implements neuromorphic vision tracking efficiently, accurately and robustly. We assume that the tracking object in the first time bin has been determined by drawing a rectangular window.

As Figure.5 shows, the algorithm mainly contains three components which are multiscale filter bank, compressive measurement matrix, and classifier. For each sample $z \in R^{w \times h}$, we convolve it with a set of rectangle filters at multiple scales to solve the scale problem. The filters $\{h_{1,1}, h_{1,2}, \cdots, h_{w,h}\}$ are defined as:

$$h_{x,y}(i,j) = \begin{cases} 1, & 0 \leq i \leq x, 0 \leq j \leq y \\ 0, & otherwise \end{cases} \quad (3)$$

where *x* and *y* are the width and height of the rectangle filter. Each filtered sample is represented as a column vector. Then we concatenate these vectors as a high-dimensional feature vector whose dimensionality is $m = (wh)^2$. The dimensionality m is about in the order of $10^6$ to $10^8$. We employ a sparse random matrix $M \in R^{n \times m}$ to project the feature vector *x* to a low-dimensional space $v \in R^n$ based on compressive sensing theory

$$v = Mx \quad (4)$$

where $n \ll m$. The random matrix for compress sensing should satisfy the Johnson-Lindenstrauss lemma[22]. The *v* preserves most of the information in *x*. Then we can classify and recognize the high-dimensional signals via their low-dimensional representation. In this paper, we adopt a typical measurement matrix, the random Gaussian matrix $M \in R^{n \times m}$ where $m_{ij} \sim N(0,1)$ which is used in many works [5, 6, 23]. The entries of the matrix are defined as

$$m_{ij} = \sqrt{s} \times \begin{cases} 1, & \text{with probability } \frac{1}{2s} \\ 0, & \text{with probability } 1 - \frac{1}{s} \\ -1, & \text{with probability } \frac{1}{2s} \end{cases} \quad (5)$$

In this paper, the $s$ is set into $m/4$, which makes a very sparse measurement matrix. Using the equation (4), we get the low-dimensional representation $v = (v_1, v_2, \cdots, v_n) \in R^n$ of each sample. Here, all the elements in v are assumed to be independently distributed and modeled with a naïve Bayes classifier[24]. We assume uniform prior as following:

$$p(y = 1) = p(y = 0) \quad (6)$$

where $y \in \{0, 1\}$ is a binary variable which represents the label of samples. Then the classifier is defined

$$H(v) = \sum_{i=1}^{n} \log\left(\frac{p(v_i|y=1)}{p(v_i|y=0)}\right), \quad (7)$$

The conditional distributions $p(v_i|y = 1)$ and $p(v_i|y = 0)$ in the above classifier are assumed to be Gaussian distribute[25] with four parameters $(\mu_i^1, \sigma_i^1, \mu_i^0, \sigma_i^0)$. The parameters are updated by

$$\mu_i^1 \leftarrow \lambda \mu_i^1 + (1 - \lambda)\mu^1 \quad (8)$$

$$\sigma_i^1 \leftarrow \sqrt{\lambda(\sigma_i^1)^2 + (1 - \lambda)(\sigma^1)^2 + \lambda(1 - \lambda)(\mu_i^1 - \mu^1)^2} \quad (9)$$

$$\mu^1 = \frac{1}{n}\sum_{k=0|y=1}^{n-1} v_i(k) \quad (10)$$

$$\sigma^1 = \sqrt{\frac{1}{n}\sum_{k=0|y=1}^{n-1}(v_i(k) - \mu^1)^2} \quad (11)$$

where $\lambda > 0$ is a learning factor. With the feature vectors, we summarize the algorithm as follows:

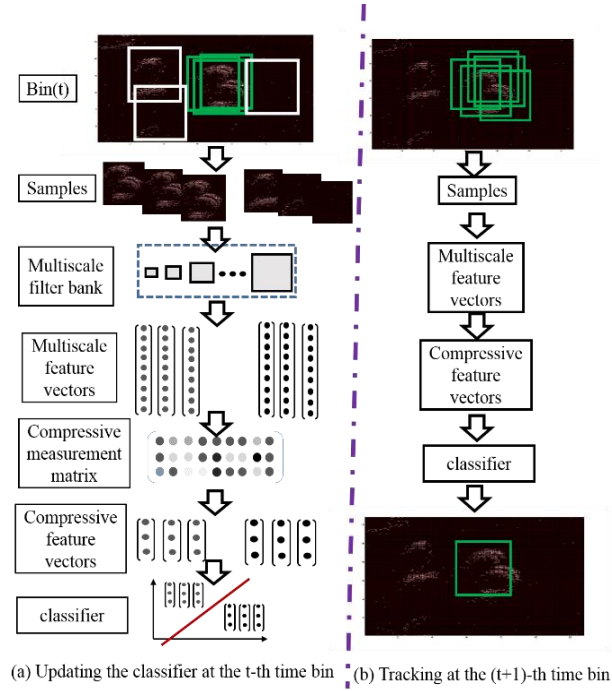

(a) Updating the classifier at the t-th time bin    (b) Tracking at the (t+1)-th time bin

Figure.5. Algorithm components of the real-time compressive tracking based on neuromorphic vision

1. In the current time bin, sample multiple rectangular windows near the tracking object, and extract the features of the samples with low dimensionality.
2. Put every feature vector of the samples to a classifier to find the maximal classifier response as the tracking location.
3. Sample both the tracking object and the background with two sets of rectangular windows, extract the low-dimensional features and update the classifier parameters as the classifier of the next time bin.

## III. RESULT

We have develop the tracking system with a silicon retina. We construct two dynamic scenes, one of which is a moving ball, while the other is a moving DVS to capture the static scene with many pictures. We capture all the spike streams in a long time window, and save them in a PC. The tracking algorithm is run on PC. For each pixel, the time resolution of spikes is 10 us. All the pixels fir spikes independently. So the DVS has a high time resolution. In the scene of a moving object, the algorithm tracks the position of the object by draws a red rectangular window, as Figure.6 and Figure.7 shows.

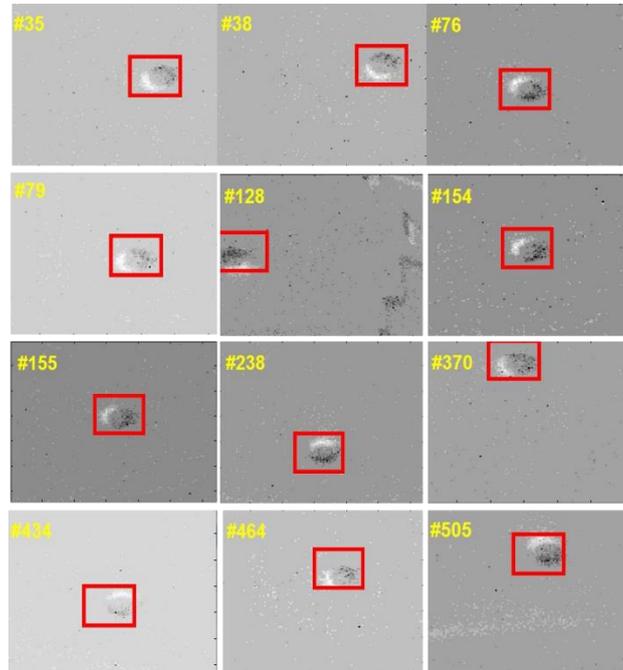

Figure.6. Real-tracking result of a moving ball. The red rectangular window represents the current positon of the ball. The yellow digit indicates the order of the time bin.

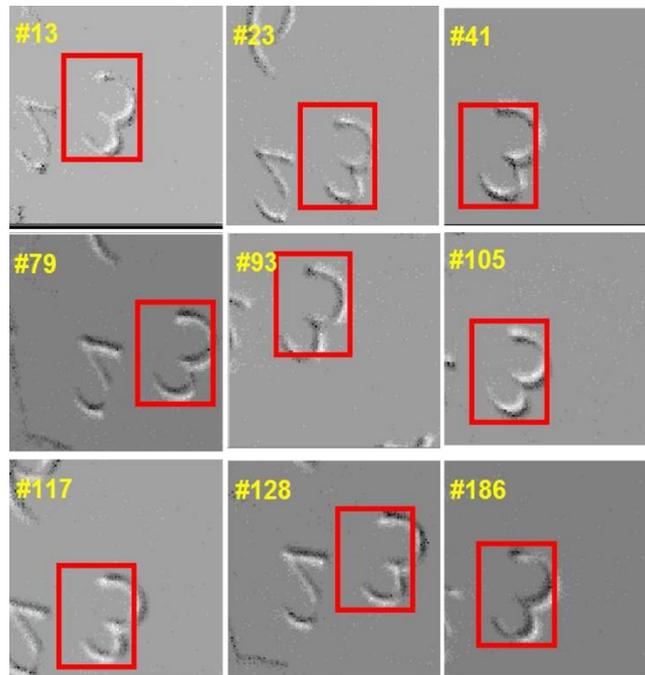

Figure.7. Real-tracking result of the digit '3' in complicated scene by moving the silicon retina. The red rectangular window represents the current positon of the digit. The yellow digit indicates the order of the time bin.

In Figure.6, the moving process of 6 s is divided into 600 time bins. In each time bin, the tracking algorithm recognizes the moving ball accurately, and shifts the red rectangular window to the new position. The ball moves at a speed of 0.5 m/s. We randomly select twelve instantaneous examples in about six hundred time bins, and the moving ball is tracked accurately with a tracking speed of 100 bins per second. In each time bin, there is an average number of about 500 events, which means that we only store and process only about 3% of the amount of data of the $128 \times 128$ pixel frame comparing to the traditional cameras. That reduces the computational requirements and the storage requirements highly.

The Figure.7 shows that the picture of digit '3' in a static scene is tracked by moving the DVS. The DVS is moving at a speed of about 0.5 m/s. The moving process of 4 s is divided into 400 bins. We select 9 instantaneous examples in the successive time bins at random, and the result shows that the algorithm does not lose the object at the 100 bins per second tracking speed. In each time bin, there is an average number of about 1000 events, which reduces the computational requirements and the storage requirements in the same way.

## IV. CONCLUSION

We have developed a real-time tracking system based on neuromorphic vision using the compressive tracking algorithm. First of all, we use the spike count coding mechanism to describe the spike streams, and demonstrate that spike count coding is suited for the silicon retina. In addition, we adapt the tracking algorithm in traditional computer vision to the neuromorphic vision and achieve a good performance, which demonstrates the method used in computer vision can be used to deal with the tasks of the neuromorphic vison. For example, as we have constructed an appearance model of the silicon retina using spike count coding in each time bin, we can adapt other more effective tracking algorithms in computer vision to the neuromorphic vision. Finally, neuromorphic vision is a vision processing mechanism which is based on silicon retinas with advantages of high time resolution, low power consumption and low data redundancy. The neuromorphic vison has broad application prospects not only in the real-time tracking, but also in the tasks of traditional computer vision. With the development of silicon retinas and algorithms, neuromorphic vision will achieve more attentions.